\documentclass[runningheads]{llncs}
\usepackage[T1]{fontenc}
\usepackage{amsmath, amssymb}
\usepackage{mathrsfs} 
\usepackage{hyperref}
\usepackage{float}
\usepackage{xcolor}
\usepackage{enumitem}
\usepackage{stmaryrd}
\usepackage{todonotes}
\usepackage{tikz}
\usepackage{tikzit}

\tikzstyle{edge}=[fill=white, draw=black, shape=circle, inner sep=1.5pt]
\tikzstyle{node}=[fill=black, draw=black, shape=circle, inner sep=1.5pt]
\tikzstyle{morphism}=[fill=white, draw=black, shape=rectangle]
\tikzstyle{negate}=[fill=red, draw=red, shape=rectangle]

\tikzstyle{pointy}=[->]
\tikzstyle{dashy}=[-, dashed]
\tikzstyle{dashpoint}=[dashed, ->]
\tikzstyle{bluefill}=[-, fill={rgb,255: red,190; green,240; blue,255}]
\tikzstyle{greyfill}=[-, fill={rgb,255: red,240; green,240; blue,240}]

\newcommand{\deftext}[1]{\textbf{#1}}
\newcommand{\defeq}[0]{\ensuremath{:=}}
\newcommand{\cp}[0]{\ensuremath{\fatsemi}} 
\newcommand{\id}[0]{\ensuremath{\mathsf{id}}}

\newcommand{\cat}[1]{\ensuremath{\mathscr{#1}}}

\newcommand{\functor}[1]{\ensuremath{\mathsf{#1}}}

\newcommand{\proj}[0]{\ensuremath{\pi}}

\newcommand{\POLY}[0]{\ensuremath{\mathsf{POLY}}}
\newcommand{\PolyCirc}[0]{\ensuremath{\mathsf{PolyCirc}}}
\newcommand{\PolyCircEq}[0]{\ensuremath{\mathsf{PolyCirc}^=}}

\newcommand{\FinSet}[0]{\ensuremath{\mathsf{FinSet}}}
\newcommand{\ToFinSet}[0]{\ensuremath{\functor{F}}}

\newcommand{\compare}[0]{\mathsf{compare}}

\newcommand{\fin}[1]{\ensuremath{\mathbf{\bar{#1}}}}
\newcommand{\set}[1]{\ensuremath{\mathbb{#1}}}
\newcommand{\Z}[0]{\set{Z}}
\newcommand{\N}[0]{\set{N}}

\newcommand{\Sat}[0]{\mathsf{Sat}}

\newcommand{\R}[0]{\ensuremath{\mathsf{R}}}
\newcommand{\Rd}[1]{\ensuremath{\R\left[\input{#1.tikz}\right]}}
\newcommand{\D}[0]{\ensuremath{\mathsf{D}}}

\newcommand{\generator}[1]{\scalebox{0.6}{\input{#1.tikz}}}

\newcommand{\Obj}[0]{\ensuremath{\mathit{Obj}}}

\begin{document}
\title{Categories of Differentiable Polynomial Circuits for Machine Learning}

\author{Paul Wilson\inst{1,2}\orcidID{0000-0003-3575-135X} \and
Fabio Zanasi\inst{2}\orcidID{0000-0001-6457-1345}}
\authorrunning{P. Wilson \& F. Zanasi}
%
\institute{University of Southampton \and University College London}

\maketitle


\begin{abstract}
  \label{section:abstract}
Reverse derivative categories (RDCs) have recently been shown to be a suitable semantic
framework for studying machine learning algorithms.
Whereas emphasis has been put on training methodologies, less attention has
been devoted to particular \emph{model classes}: the concrete categories whose
morphisms represent machine learning models.
In this paper we study presentations by generators and equations of classes of RDCs. In particular, we propose \emph{polynomial circuits} as a suitable machine learning model. We give an axiomatisation for these circuits and prove a functional completeness result. Finally, we discuss the use of polynomial circuits over specific semirings to perform
machine learning with discrete values.

\end{abstract}

\section{Introduction}
\label{section:introduction}
Reverse Derivative Categories~\cite{rdc} have recently been introduced as a
formalism to study abstractly the concept of differentiable functions.
As explored in~\cite{cfgbl}, it turns out that this framework is
suitable to give a categorical semantics for gradient-based learning.
In this approach, models--as for instance neural networks--correspond to
morphisms in some RDC.
We think of the particular RDC as a `model class'--the space of all possible
definable models.

However, much less attention has been directed to actually defining the
RDCs in which models are specified: existing approaches assume there is some
chosen RDC and morphism, treating both essentially as a black box.
In this paper, we focus on classes of RDCs which we call `polynomial circuits',
which may be thought of as a more expressive version of the boolean circuits of Lafont~\cite{lafont},
with wires carrying values from an arbitrary semiring instead of $\Z_2$.
Because we ensure polynomial circuits have RDC structure, they are suitable as machine learning models, as we discuss in the second part of the
paper.

Our main contribution is to provide an algebraic description of polynomial circuits and their reverse derivative structure. More specifically, we build a presentation of these categories by operation and equations.
Our approach will proceed in steps, by gradually enriching the algebraic structures considered, and culminate in showing that a certain presentation is
\emph{functionally complete} for the class of functions that these circuits are meant
to represent.

An important feature of our categories of circuits is that morphisms are specified in the graphical formalism of
\emph{string diagrams}.
This approach has the benefit of making the model specification reflect
its combinatorial structure.
Moreover, at a computational level, the use of string diagrams makes available
the principled mathematical toolbox of \emph{double-pushout rewriting}, via an
interpretation of string diagrams as hypergraphs~\cite{sdr1,sdr2,sdr3}.
Finally, the string diagrammatic presentation suggests a way to encode
polynomial circuits into datastructures: an important requirement for being able
to incorporate these models into tools analogous to existing deep learning
frameworks such as TensorFlow~\cite{tensorflow} and PyTorch~\cite{pytorch}.

Tool-building is not the only application of the model classes we define here.
Recent neural networks literature~\cite{bengio2013estimating,ibm_pact}
proposes to improve model performance (e.g. memory
requirements, power consumption, and inference time) by `quantizing' network
parameters.
One categorical approach in this area is \cite{rda}, in which the authors define
learning directly over boolean circuit models instead of training with
real-valued parameters and then quantizing.
The categories in our paper can be thought of as a generalisation of this
approach to arbitrary semirings.

This generalisation further yields another benefit: while
neural networks literature focuses on finding particular `architectures' (i.e. specific morphisms)
that work well for a given problem,
our approach suggests a new avenue for model design:
changing the underlying semiring (and thus the corresponding notion of
arithmetic).
To this end, we conclude our paper with some examples of finite semirings which may
yield new approaches to model design.


\paragraph{Synopsis} We recall the notion of RDC in Section \ref{section:rdcs}, and then
study presentations of RDCs by operations and equations in Section
\ref{section:extension-theorem}.
We define categories of polynomial circuits in Section
\ref{section:polynomial-circuits}, before showing how they can be made
\emph{functionally complete} in Section \ref{section:functional-completeness}.
Finally, we close by discussing some case studies of polynomial circuits in machine learning, in Section
\ref{section:saturating-arithmetic}.

\section{Reverse Derivative Categories}
\label{section:rdcs}
We recall the notion of reverse derivative category~\cite{rdc} in two steps. First we introduce the simpler structure of cartesian left-additive categories. We make use of the graphical formalism of \emph{string diagrams}~\cite{selinger} to represent morphisms in our categories.
\begin{definition}\label{def:cartesianleftadditive}
  A \deftext{Cartesian Left-Additive Category} (\cite{rdc}, \cite{blute_cartesian_2009})
  is a cartesian category in which each object $A$ is equipped with a commutative monoid and zero map:
\begin{equation}\label{eq:genmonoid}
 \input{g-obj-add.tikz} \qquad \qquad \begin{tikzpicture}
	\begin{pgfonlayer}{nodelayer}
		\node [style=node] (0) at (0, 0) {};
		\node [style=none] (1) at (0.75, 0) {};
		\node [style=none] (6) at (1.5, 0) {$A$};
	\end{pgfonlayer}
	\begin{pgfonlayer}{edgelayer}
		\draw (1.center) to (0);
	\end{pgfonlayer}
\end{tikzpicture}

 \end{equation}
  so that
  \begin{align}
    \label{equation:cla}
    \begin{split}
      \input{s-addtwist.tikz} = \input{g-add.tikz} \qquad
      \input{cla-assoc-lhs.tikz} = \input{cla-assoc-rhs.tikz} \qquad
      \input{cla-unit-lhs.tikz} = \begin{tikzpicture}
	\begin{pgfonlayer}{nodelayer}
		\node [style=none] (1) at (0.5, 0) {};
		\node [style=none] (2) at (-0.5, 0) {};
	\end{pgfonlayer}
	\begin{pgfonlayer}{edgelayer}
		\draw (2.center) to (1.center);
	\end{pgfonlayer}
\end{tikzpicture}

      \\
      \input{cla-coherence-a-lhs.tikz} \enspace = \enspace \input{cla-coherence-a-rhs.tikz} \qquad \qquad
      \begin{tikzpicture}
	\begin{pgfonlayer}{nodelayer}
		\node [style=node] (0) at (-0.75, 0) {};
		\node [style=none] (1) at (0, 0) {};
		\node [style=none] (2) at (1, 0) {$A \otimes B$};
	\end{pgfonlayer}
	\begin{pgfonlayer}{edgelayer}
		\draw (0) to (1.center);
	\end{pgfonlayer}
\end{tikzpicture}
 \enspace = \enspace \input{cla-coherence-z-rhs.tikz}
    \end{split}
  \end{align}
\end{definition}

Note that the category being cartesian means that: (I) it is symmetric monoidal,
namely for each object $A$ and $B$ there are symmetries
\generator{g-obj-twist}
and identities
\generator{g-obj-identity}
satisfying the laws of symmetric monoidal
categories~\cite{selinger}; (II) each object $A$  comes equipped with a
\emph{copy} and a \emph{discard} map:
\begin{equation}\label{eq:gencomonoid}
	 \input{g-obj-copy.tikz} \qquad \begin{tikzpicture}
	\begin{pgfonlayer}{nodelayer}
		\node [style=node] (0) at (0, 0) {};
		\node [style=none] (1) at (-0.75, 0) {};
		\node [style=none] (6) at (-1.5, 0) {$A$};
	\end{pgfonlayer}
	\begin{pgfonlayer}{edgelayer}
		\draw (1.center) to (0);
	\end{pgfonlayer}
\end{tikzpicture}

\end{equation}
satisfying the axioms of commutative comonoids and natural with respect to the other morphisms in the category:
\begin{align}
  \label{equation:cartesian}
  \begin{split}
    \input{s-copytwist.tikz} = \input{g-copy.tikz} \qquad
    \input{fox-assoc-lhs.tikz} = \input{fox-assoc-rhs.tikz} \qquad
    \input{fox-counit.tikz} = \begin{tikzpicture}
	\begin{pgfonlayer}{nodelayer}
		\node [style=none] (1) at (0.5, 0) {};
		\node [style=none] (2) at (-0.5, 0) {};
	\end{pgfonlayer}
	\begin{pgfonlayer}{edgelayer}
		\draw (2.center) to (1.center);
	\end{pgfonlayer}
\end{tikzpicture}
 \qquad
    \\
    \input{fox-copy-naturality-lhs.tikz} = \input{fox-copy-naturality-rhs.tikz}
    \qquad \qquad
    \input{fox-discard-naturality-lhs.tikz} = \begin{tikzpicture}
	\begin{pgfonlayer}{nodelayer}
		\node [style=node] (8) at (0, 0) {};
		\node [style=none] (9) at (-0.75, 0) {};
	\end{pgfonlayer}
	\begin{pgfonlayer}{edgelayer}
		\draw (9.center) to (8);
	\end{pgfonlayer}
\end{tikzpicture}

  \end{split}
\end{align}

\begin{remark}
Definition~\ref{def:cartesianleftadditive} is given differently than the standard definition of cartesian left-additive categories~\cite[Definition 1]{rdc}, which one may recover by letting addition of morphisms be $f + g :=
  \scalebox{0.6}{\input{cla-morphism-addition.tikz}}$, and the zero morphism be $0 :=
  \scalebox{0.6}{\begin{tikzpicture}
	\begin{pgfonlayer}{nodelayer}
		\node [style=node] (13) at (-0.5, 0) {};
		\node [style=none] (14) at (-1.25, 0) {};
		\node [style=node] (31) at (0.5, 0) {};
		\node [style=none] (32) at (1.25, 0) {};
	\end{pgfonlayer}
	\begin{pgfonlayer}{edgelayer}
		\draw (14.center) to (13);
		\draw (32.center) to (31);
	\end{pgfonlayer}
\end{tikzpicture}
}$.
Equations of cartesian left-additive categories as given in~\cite[Definition 1]{rdc}
  \begin{align*}
     x \cp (f + g) \ = \ (x \cp f) + (x \cp g) \qquad \qquad
     x \cp 0 \ = \ 0
  \end{align*}
are represented by string diagrams
  \begin{align*}
    \input{cla-axiom-left-additivity-lhs.tikz} \quad = \quad \input{cla-axiom-left-additivity-rhs.tikz}
    \qquad \qquad
    \input{cla-axiom-left-additivity-2-lhs.tikz} \quad = \quad \begin{tikzpicture}
	\begin{pgfonlayer}{nodelayer}
		\node [style=node] (13) at (0.5, 0) {};
		\node [style=none] (43) at (-0.5, 0) {};
	\end{pgfonlayer}
	\begin{pgfonlayer}{edgelayer}
		\draw (43.center) to (13);
	\end{pgfonlayer}
\end{tikzpicture}

  \end{align*}
  and follow from Definition~\ref{def:cartesianleftadditive} thanks to the naturality of
  $\input{g-copy.tikz}$ and $\begin{tikzpicture}
	\begin{pgfonlayer}{nodelayer}
		\node [style=node] (0) at (0, 0) {};
		\node [style=none] (1) at (-0.75, 0) {};
	\end{pgfonlayer}
	\begin{pgfonlayer}{edgelayer}
		\draw (1.center) to (0);
	\end{pgfonlayer}
\end{tikzpicture}
$, respectively.
  We refer to~\cite[Proposition 1.2.2 (iv)]{blute_cartesian_2009} for more
  details on the equivalence of the two definitions.
\end{remark}

Now, Reverse Derivative Categories, originally defined in \cite{rdc}, are
cartesian left-additive categories equipped with an operator $\R$ of the
following type, and satisfying axioms RD.1 - RD.7 detailed in \cite[Definition 13]{rdc}.

\[ \frac{A \overset{f}{\longrightarrow} B}{A \times B \underset{\R[f]}{\longrightarrow} A} \]

Intuitively, for a morphism $f : A \to B$ we think of its reverse derivative $R[f] : A
\times B \to A$ as approximately computing the change of \emph{input} to $f$
required to achieve a given change in \emph{output}.  That is, if $f$ is a
function, we should have

\[ f(x) + \delta_y \approx f(x + \R[f](x, \delta_y)) \]

The authors of \cite{rdc} go on to show that any reverse derivative category
also admits a \emph{forward} differential structure: i.e, it is also a Cartesian
Differential Category (CDC).
This means the existence of a forward differential operator $\D$ satisfying various axioms,
and having the following type:

\[ \frac{A \overset{f}{\longrightarrow} B}{A \times A \underset{\D[f]}{\longrightarrow} B} \]

In an RDC, the forward differential operator is defined in terms of $\R$ as
the following string diagram, with $\R^{(n)}$ denoting the $n$-fold application\footnote{
  For example, $R^{(2)}[f]$ denotes the map $R[R[f]]$.
} of $\R$:

\[ \D[f] \quad \defeq \quad \input{rdc-definition-noadapters-d-rhs.tikz} \]

In contrast to the $\R$ operator, we think of $\D$ as computing a change in
\emph{output} from a given change in \emph{input}, whence `forward' and
`reverse' derivative:

\[ f(x + \delta_x) \approx f(x) + \D[f](x, \delta_x) \]

The final pieces we need to state our definition of RDCs are the (cartesian differential) notions of
\emph{partial derivative} and \emph{linearity} defined in \cite{rdc}.
Graphically, the partial derivative of $g : A \times B \to C$ with respect to
$B$ is defined as follows:

\[ \D_B[g] \quad \defeq \quad \input{rdc-definition-noadapters-dB-rhs.tikz} \]

Finally we say that $g$ is \emph{linear in $B$} when
\[ \D_B[g] \quad = \quad \input{rdc-definition-linearB-rhs.tikz} \]
and more generally that $f : A \to B$ is \emph{linear} when
\[ \D[f] \quad = \quad \input{rdc-definition-linear-rhs.tikz} \]


We can now formulate the definition of RDCs. Note that in the following
definition and proofs we treat $\D$ purely as a syntactic shorthand for its
definition in terms of $\R$.
We avoid use of CDC axioms to prevent a circular definition, although one can
derive them as corollaries of the RDC axioms.

\begin{definition}
  \label{definition:rdc}
  A \deftext{Reverse Derivative Category} is a cartesian left-additive category
  equipped with a \emph{reverse differential combinator} $\R$:
  \[ \frac{A \overset{f}{\longrightarrow} B}{A \times B \underset{\R[f]}{\longrightarrow} A} \]
	satisfying the following axioms: \\
  \underline{\textbf{[ARD.1]}} (Structural axioms, equivalent to RD.1, RD.3-5 in \cite{rdc}) \\
  \begin{align*}
    & \R\left[\begin{tikzpicture}
	\begin{pgfonlayer}{nodelayer}
		\node [style=none] (1) at (0.5, 0) {};
		\node [style=none] (2) at (-0.5, 0) {};
	\end{pgfonlayer}
	\begin{pgfonlayer}{edgelayer}
		\draw (2.center) to (1.center);
	\end{pgfonlayer}
\end{tikzpicture}
\right] = \begin{tikzpicture}
	\begin{pgfonlayer}{nodelayer}
		\node [style=node] (13) at (0.5, 0.25) {};
		\node [style=none] (43) at (-0.5, 0.25) {};
		\node [style=none] (44) at (-0.5, -0.25) {};
		\node [style=none] (45) at (1, -0.25) {};
	\end{pgfonlayer}
	\begin{pgfonlayer}{edgelayer}
		\draw (43.center) to (13);
		\draw (44.center) to (45.center);
	\end{pgfonlayer}
\end{tikzpicture}

    & \R\left[\input{g-copy.tikz}\right]     & = \input{r-copy.tikz}
    & \R\left[\input{g-add.tikz}\right]      & = \input{r-add.tikz}
    \\
    & \R\left[\input{g-twist.tikz}\right]    = \input{r-twist.tikz}
    & \R\left[\begin{tikzpicture}
	\begin{pgfonlayer}{nodelayer}
		\node [style=node] (0) at (0, 0) {};
		\node [style=none] (1) at (-0.75, 0) {};
	\end{pgfonlayer}
	\begin{pgfonlayer}{edgelayer}
		\draw (1.center) to (0);
	\end{pgfonlayer}
\end{tikzpicture}
\right]  & = \begin{tikzpicture}
	\begin{pgfonlayer}{nodelayer}
		\node [style=node] (13) at (-0.5, 0) {};
		\node [style=none] (14) at (-1.25, 0) {};
		\node [style=node] (31) at (0.5, 0) {};
		\node [style=none] (32) at (1.25, 0) {};
	\end{pgfonlayer}
	\begin{pgfonlayer}{edgelayer}
		\draw (14.center) to (13);
		\draw (32.center) to (31);
	\end{pgfonlayer}
\end{tikzpicture}

    & \R\left[\begin{tikzpicture}
	\begin{pgfonlayer}{nodelayer}
		\node [style=node] (0) at (0, 0) {};
		\node [style=none] (1) at (0.75, 0) {};
	\end{pgfonlayer}
	\begin{pgfonlayer}{edgelayer}
		\draw (1.center) to (0);
	\end{pgfonlayer}
\end{tikzpicture}
\right]     & = \begin{tikzpicture}
	\begin{pgfonlayer}{nodelayer}
		\node [style=node] (0) at (0, 0) {};
		\node [style=none] (1) at (-0.75, 0) {};
	\end{pgfonlayer}
	\begin{pgfonlayer}{edgelayer}
		\draw (1.center) to (0);
	\end{pgfonlayer}
\end{tikzpicture}

  \end{align*}
  \[
    \R[f \cp g] = \input{r-composition.tikz}
    \qquad \qquad
    \R[f \times g] = \input{r-tensor.tikz}
  \]
  \underline{\textbf{[ARD.2]}} (Additivity of change, equivalent to RD.2 in \cite{rdc}) \\
  \[
    \input{rdc-rd2-a-lhs.tikz} = \input{rdc-rd2-a-rhs.tikz}
    \qquad \qquad
    \input{rdc-rd2-b-lhs.tikz} = \begin{tikzpicture}
	\begin{pgfonlayer}{nodelayer}
		\node [style=node] (13) at (-0.5, 0) {};
		\node [style=none] (14) at (-1.25, 0) {};
		\node [style=node] (31) at (0.5, 0) {};
		\node [style=none] (32) at (1.25, 0) {};
	\end{pgfonlayer}
	\begin{pgfonlayer}{edgelayer}
		\draw (14.center) to (13);
		\draw (32.center) to (31);
	\end{pgfonlayer}
\end{tikzpicture}

  \]
  \underline{\textbf{[ARD.3]}} (Linearity of change, equivalent to RD.6 in \cite{rdc}) \\
	\[ \D_B\left[\R[f]\right] = \input{rdc-rd6-rhs.tikz} \]
  \underline{\textbf{[ARD.4]}} (Symmetry of partials, equivalent to RD.7 in \cite{rdc}) \\
	\[ \D^{(2)}[f] = \input{rdc-rd7-alt-rhs.tikz} \]
\end{definition}

\begin{remark}
  Note that we may alternatively write axioms ARD.3 and ARD.4 directly in terms
  of the $\R$ operator by simply expanding the syntactic definition of $\D$.
\end{remark}

Note that axioms ARD.1 and ARD.2 are quite different to that of \cite{rdc},
while ARD.3 and ARD.4 are essentially direct restatements in graphical language
of RD.6 and RD.7 respectively.

The definition we provide best suits our purposes, although it is different than the standard one provided in \cite[Definition 13]{rdc}. We can readily verify that they are equivalent.

\begin{theorem}
  Definition \ref{definition:rdc} is equivalent to \cite[Definition 13]{rdc}.
\end{theorem}

\begin{proof}
  Axioms ARD.3-4 are direct statements of axioms RD.6-7, so it suffices to show
  that we can derive axioms ARD.1-2 from RD.1.5 and vice-versa.
  The structural axioms ARD.1 follow directly from RD.1 and RD.3-5.
  \begin{itemize}
    \item For $\R\left[\generator{g-identity}\right]$ use RD.3 directly.
    \item For $\R\left[\generator{g-twist}\right]$, apply RD.4 to $\langle\proj_1, \proj_0\rangle$
    \item For $\R\left[\generator{g-add}\right]$, apply RD.1 to $\proj_0 + \proj_1$
    \item For $\R\left[\generator{g-zero}\right]$, apply RD.1 directly.
    \item For $\R\left[\generator{g-copy}\right]$, apply RD.4 to $\langle \id, \id \rangle$
    \item For $\R\left[\generator{g-discard}\right]$, apply RD.4 directly.
    \item For composition $f \cp g$, apply RD.5 directly
    \item For tensor $f \times g$, apply RD.4 to $\langle \proj_0 \cp f, \proj_1 \cp g \rangle$
  \end{itemize}
  In the reverse direction, we can obtain RD.1 and RD.3-5 by simply constructing
  each equation and showing it holds given the structural equations. For
  example, RD.1 says that $\R[f + g] = \R[f] + \R[g]$ and $\R[0] = 0$, which we can write graphically as:
  \[ \R\left[\input{cla-morphism-addition.tikz}\right] = \R\left[f\right] + \R\left[g\right] \]
  and
  \[ \R\left[\begin{tikzpicture}
	\begin{pgfonlayer}{nodelayer}
		\node [style=node] (13) at (-0.5, 0) {};
		\node [style=none] (14) at (-1.25, 0) {};
		\node [style=node] (31) at (0.5, 0) {};
		\node [style=none] (32) at (1.25, 0) {};
	\end{pgfonlayer}
	\begin{pgfonlayer}{edgelayer}
		\draw (14.center) to (13);
		\draw (32.center) to (31);
	\end{pgfonlayer}
\end{tikzpicture}
\right] = \input{r-morphism-zero.tikz} \]
  ARD.2 can be derived from RD.2 by setting $a$, $b$, $c$ to appropriate projections,
  and in the reverse direction we can obtain RD.2 simply
  by applying ARD.2 to its left-hand-side and using naturality of $\generator{g-copy}$.
\end{proof}

A main reason to give an alternative formulation of cartesian left-additive and
reverse derivative categories is being able to work with a more `algebraic'
definition, which revolves around the interplay of operations
$\generator{g-add}$,
$\generator{g-zero}$,
$\generator{g-copy}$,
and
$\generator{g-discard}$.
This perspective is particularly useful when one wants to show that the free
category on certain generators and equations has RDC structure.
We thus recall such free construction, referring to~\cite[Chapter
2]{fabio_thesis} and \cite[Section 5]{BaezNetworkTheory} for a more thorough exposition.

\begin{definition}
	Given a set $\Obj$ of \emph{generating objects}, we may consider a set $\Sigma$ of \emph{generating morphisms} $f \colon w \to v$, where the \emph{arity} $w \in \Obj^{\star}$ and the \emph{coarity} $v \in \Obj^{\star}$ of $f$ are $\Obj$-words. Cartesian left-additive \emph{$\Sigma$-terms} are defined inductively:
	\begin{itemize}
		\item Each $f \colon w \to v$ is a $\Sigma$-term.
		\item For each $A \in \Obj$, the generators~\eqref{eq:genmonoid} and \eqref{eq:gencomonoid} of the cartesian left-additive structure are $\Sigma$-terms.
		\item If $f \colon w \to v$, $g \colon v \to u$, and $h \colon w' \to v'$ are $\Sigma$-terms, then $f \cp g \colon w \to u$ and $f \otimes h \colon ww' \to vv'$ are $\Sigma$-terms, represented as string diagrams
		\begin{equation*}
      \input{free-composition.tikz} \qquad \input{free-tensor.tikz}
		\end{equation*}
	\end{itemize}
  Let us fix $\Obj$, $\Sigma$ and a set $E$ of equations between $\Sigma$-terms.
  The cartesian left-additive category $\cat{C}$ freely generated by
  $(\Obj, \Sigma, E)$
  is the monoidal category with set of objects $\Obj^{\star}$ and
  morphisms the $\Sigma$-terms quotiented by the axioms of cartesian
  left-additive categories and the equations in $E$. The monoidal product in
  $\cat{C}$ is given on objects by word concatenation. Identities, monoidal
  product and sequential composition of morphisms are given by the corresponding
  $\Sigma$-terms and their constructors $f \otimes h$ and $f \cp g$.
\end{definition}

One may readily see that $\cat{C}$ defined in this way is indeed cartesian
left-additive.
We say that $\cat{C}$ is \emph{presented} by generators
$(\Obj, \Sigma)$ and equations $E$.

\section{Reverse Derivatives and Algebraic Presentations}
\label{section:extension-theorem}
As we will see in Section~\ref{section:functional-completeness}, our argument
for functional completeness relies on augmenting the algebraic presentation of
polynomial circuits with an additional operation. To formulate such result, we
first need to better understand how reverse differential combinators may be
defined compatibly with the generators and equations presenting a category.



\begin{theorem}
  \label{theorem:extension}
  Let $\cat{C}$ be the cartesian left-additive category presented by generators
  $(\Obj,\Sigma)$ and equations $E$.
  If for each $s \in \Sigma$ there is some $\R[s]$ which is well-defined
  (see Remark \ref{remark:well-defined}) with respect to $E$,
  and which satisfies axioms ARD.1-4,
  then $\cat{C}$ is a reverse derivative category.
\end{theorem}

\begin{proof}
  Observe that axioms ARD.1 fix the definition of $\R$ on composition, tensor
  product and the cartesian and left-additive structures.
  It therefore suffices to show that axioms ARD.2-4 are preserved by composition
  and tensor product.
  That is, for morphisms $f, g$ of appropriate types, both $f \cp g$ and $f
  \otimes g$ preserve axioms ARD.2-4.
  Thus, any morphism constructed from generators must also satisfy the axioms ARD.1-4,
  and $\cat{C}$ must be an RDC.
  We provide the full graphical proofs that ARD.2-4 are preserved by composition
  and tensor product in Appendix~\ref{section:appendix-extension-theorem}.
\end{proof}

\begin{remark}
  \label{remark:well-defined}
	In the statement of Theorem~\ref{theorem:extension}, strictly speaking $s \in \Sigma$ is just a representative of the equivalence class of $\Sigma$-terms (modulo $E$ plus the laws of left-additive cartesian categories) defining a morphism in $\cat{C}$. Because of this, we require $\R[s]$ to be `well-defined', in the sense that if $s$ and $t$ are representatives of the same morphisms of $\cat{C}$, then the same should hold for $\R[s]$ and $\R[t]$. In a nutshell, we are allowed to define $\R$ directly on $\Sigma$-terms, provided our definition is compatible with $E$ and the laws of left-additive cartesian categories.
\end{remark}

An immediate consequence of Theorem \ref{theorem:extension} is that if we have a
presentation of an RDC $\cat{C}$, we can `freely extend' it with an additional
operation $s$, a chosen reverse derivative $\R[s]$, and equations $E'$, so long as
$\R$ is well-defined with respect to $E'$ and the axioms ARD.2-4 hold for
$\R[s]$.
Essentially, this gives us a simple recipe for adding new `gadgets' to existing
RDCs and ensuring they retain RDC structure.

One particularly useful such `extension' is the addition of a
\emph{multiplication} morphism $\generator{g-multiply}$ that distributes over
the addition $\generator{g-add}$.
We define categories with such a morphism as an extension of cartesian
left-additive categories as follows:

\begin{definition}
  \label{definition:cartesian-distributive}
  A \deftext{Cartesian Distributive Category} is a cartesian left-additive
  category such that each object $A$ is equipped with a commutative monoid
  $\generator{g-multiply}$ and unit $\generator{g-one}$ which distributes over
  the addition $\generator{g-add}$.
  More completely, it is a category having generators
  \[
    \input{g-copy.tikz} \qquad \begin{tikzpicture}
	\begin{pgfonlayer}{nodelayer}
		\node [style=node] (0) at (0, 0) {};
		\node [style=none] (1) at (-0.75, 0) {};
	\end{pgfonlayer}
	\begin{pgfonlayer}{edgelayer}
		\draw (1.center) to (0);
	\end{pgfonlayer}
\end{tikzpicture}
 \qquad
    \input{g-add.tikz} \qquad \begin{tikzpicture}
	\begin{pgfonlayer}{nodelayer}
		\node [style=node] (0) at (0, 0) {};
		\node [style=none] (1) at (0.75, 0) {};
	\end{pgfonlayer}
	\begin{pgfonlayer}{edgelayer}
		\draw (1.center) to (0);
	\end{pgfonlayer}
\end{tikzpicture}
 \qquad
    \input{g-multiply.tikz} \qquad \begin{tikzpicture}
	\begin{pgfonlayer}{nodelayer}
		\node [style=edge] (0) at (0, 0) {};
		\node [style=none] (1) at (0.75, 0) {};
	\end{pgfonlayer}
	\begin{pgfonlayer}{edgelayer}
		\draw (1.center) to (0);
	\end{pgfonlayer}
\end{tikzpicture}
 \qquad
  \]
  satisfying the \emph{cartesianity} equations~\eqref{equation:cartesian},
  the \emph{left-additivity} equations~\eqref{equation:cla},
  the \emph{multiplicativity} equations
  \begin{equation}\label{eq:multiplicative}
    \input{s-multiplytwist.tikz} = \input{g-multiply.tikz} \qquad
    \input{multiply-assoc-lhs.tikz} = \input{multiply-assoc-rhs.tikz} \qquad
    \input{multiply-unit-lhs.tikz} = \begin{tikzpicture}
	\begin{pgfonlayer}{nodelayer}
		\node [style=none] (1) at (0.5, 0) {};
		\node [style=none] (2) at (-0.5, 0) {};
	\end{pgfonlayer}
	\begin{pgfonlayer}{edgelayer}
		\draw (2.center) to (1.center);
	\end{pgfonlayer}
\end{tikzpicture}
 \qquad
  \end{equation}
  and the \emph{distributivity} and \emph{annihilation} equations
  \begin{equation}\label{eq:distributive}
    \input{distributivity-lhs.tikz} = \input{distributivity-rhs.tikz}
    \qquad
    \input{annihilation-lhs.tikz} = \begin{tikzpicture}
	\begin{pgfonlayer}{nodelayer}
		\node [style=node] (13) at (-0.5, 0) {};
		\node [style=none] (14) at (-1.25, 0) {};
		\node [style=node] (31) at (0.5, 0) {};
		\node [style=none] (32) at (1.25, 0) {};
	\end{pgfonlayer}
	\begin{pgfonlayer}{edgelayer}
		\draw (14.center) to (13);
		\draw (32.center) to (31);
	\end{pgfonlayer}
\end{tikzpicture}

  \end{equation}
\end{definition}

Just as for cartesian left-additive categories, one may construct cartesian distributive categories freely from a set of objects $\Obj$, a signature $\Sigma$, and equations $E$, the difference being that $\Sigma$-term will be constructed using also $\generator{g-multiply}$ and $\generator{g-one}$, and quotiented also by~\eqref{eq:multiplicative}-\eqref{eq:distributive}. The main example of cartesian distributive categories are \emph{Polynomial Circuits}, which we
define in Section \ref{section:polynomial-circuits} below. 

Reverse derivative categories define a reverse differential combinator  on a left-additive cartesian structure. As cartesian distributive categories properly extend left-additive ones, it is natural to ask how we may extend the definition of the reverse differential combinator to cover the extra operations $\generator{g-multiply}$ and $\generator{g-one}$.
The following theorem provide a recipe, which we will use in the next section to study RDCs with a cartesian distributive structure.
Note that the definition of $\R^*$ below is a string diagrammatic version of the
reverse derivative combinator defined on $\POLY$ in \cite{rdc}.

\begin{theorem}
  \label{theorem:cartesian-distributive}
  Suppose $\cat{C}$ is a left-additive cartesian category presented by $(\Obj, \Sigma, E)$, and assume $\cat{C}$ is also an RDC, say with reverse differential combinator $R$. Then the cartesian distributive category $\cat{C}^*$ presented by $(\Obj, \Sigma, E)$, with reverse differential combinator $R^*$ defined as $R$ on the left-additive cartesian structure, and as follows
  \begin{equation}\label{eq:defRcartesiandistr}
    R^*\left[\input{g-multiply.tikz}\right] = \input{r-multiply.tikz}
    \qquad \qquad
    R^*\left[\begin{tikzpicture}
	\begin{pgfonlayer}{nodelayer}
		\node [style=edge] (0) at (0, 0) {};
		\node [style=none] (1) at (0.75, 0) {};
	\end{pgfonlayer}
	\begin{pgfonlayer}{edgelayer}
		\draw (1.center) to (0);
	\end{pgfonlayer}
\end{tikzpicture}
\right] = \begin{tikzpicture}
	\begin{pgfonlayer}{nodelayer}
		\node [style=node] (0) at (0, 0) {};
		\node [style=none] (1) at (-0.75, 0) {};
	\end{pgfonlayer}
	\begin{pgfonlayer}{edgelayer}
		\draw (1.center) to (0);
	\end{pgfonlayer}
\end{tikzpicture}

  \end{equation}
  on the extra distributive structure, is also an RDC.
\end{theorem}

\begin{proof}
  It suffices to check that $\R$ is well-defined with respect to the additional
  equations of cartesian distributive categories,
  and that the new generators $\generator{g-multiply}$ and $\generator{g-one}$
  satisfy axioms ARD.2-4.
\end{proof}




\section{Polynomial Circuits}
\label{section:polynomial-circuits}
Our motivating example of cartesian distributive categories is that of
\emph{polynomial circuits},
whose morphisms can be thought of as representing polynomials over a
commutative semiring.
We define them as follows:

\begin{definition}
  \label{definition:polycirc-s}
  Let $S$ be a commutative semiring.
  We define $\PolyCirc_S$ as the cartesian distributive category presented by
  (I) one generating object $1$,
  (II) for each $s \in S$, a generating morphism $\input{constant-s.tikz} \colon 0 \to 1$,
  (III) the `constant' equations
  \begin{equation}
    \label{equation:constant}
    \input{constant-zero.tikz} = \begin{tikzpicture}
	\begin{pgfonlayer}{nodelayer}
		\node [style=node] (0) at (0, 0) {};
		\node [style=none] (1) at (0.75, 0) {};
	\end{pgfonlayer}
	\begin{pgfonlayer}{edgelayer}
		\draw (1.center) to (0);
	\end{pgfonlayer}
\end{tikzpicture}

    \qquad
    \input{constant-add-lhs.tikz} = \input{constant-add-rhs.tikz}
    \qquad
    \input{constant-one.tikz} = \begin{tikzpicture}
	\begin{pgfonlayer}{nodelayer}
		\node [style=edge] (0) at (0, 0) {};
		\node [style=none] (1) at (0.75, 0) {};
	\end{pgfonlayer}
	\begin{pgfonlayer}{edgelayer}
		\draw (1.center) to (0);
	\end{pgfonlayer}
\end{tikzpicture}

    \qquad
    \input{constant-multiply-lhs.tikz} = \input{constant-multiply-rhs.tikz}
  \end{equation}
for $s,t \in S$, intuitively saying that the generating morphisms respect addition and multiplication of $S$.
\end{definition}

\begin{proposition}
  \label{proposition:polycirc-is-an-rdc}
  $\PolyCirc_S$ is an RDC with
  $\R\left[\input{constant-s.tikz}\right] = \begin{tikzpicture}
	\begin{pgfonlayer}{nodelayer}
		\node [style=node] (0) at (0, 0) {};
		\node [style=none] (1) at (-0.75, 0) {};
	\end{pgfonlayer}
	\begin{pgfonlayer}{edgelayer}
		\draw (1.center) to (0);
	\end{pgfonlayer}
\end{tikzpicture}
\ $.
\end{proposition}

\begin{proof}
  The type of $\R\left[\input{constant-s.tikz}\right] : 1 \to 0$
  implies that there is only one choice of reverse derivative, namely the unique discard map
  \generator{g-discard}.
  Furthermore, $\R$ is well-defined with respect to the constant equations
  \eqref{equation:constant} for the same reason.
  Finally, observe that the axioms ARD.2-4 hold for
  $\R\left[\input{constant-s.tikz}\right]$,
  precisely in the same way as for $\R\left[\begin{tikzpicture}
	\begin{pgfonlayer}{nodelayer}
		\node [style=node] (0) at (0, 0) {};
		\node [style=none] (1) at (0.75, 0) {};
	\end{pgfonlayer}
	\begin{pgfonlayer}{edgelayer}
		\draw (1.center) to (0);
	\end{pgfonlayer}
\end{tikzpicture}
\right]$, and so
  $\PolyCirc_S$ is an RDC.
\end{proof}

Although our Definition \ref{definition:polycirc-s} of $\PolyCirc_S$ requires
that we add an axiom for each possible addition and multiplication of constants,
for some significant choices of $S$ an equivalent smaller finite axiomatisation is possible.
We demonstrate this with some examples.

\begin{example}
  \label{example:polycirc-z2}
In the case of $\PolyCirc_{\Z_2}$, the equations of Definition~\ref{definition:polycirc-s} reduce to the single equation
  \[ \input{s-x-plus-x.tikz} = \begin{tikzpicture}
	\begin{pgfonlayer}{nodelayer}
		\node [style=node] (13) at (-0.5, 0) {};
		\node [style=none] (14) at (-1.25, 0) {};
		\node [style=node] (31) at (0.5, 0) {};
		\node [style=none] (32) at (1.25, 0) {};
	\end{pgfonlayer}
	\begin{pgfonlayer}{edgelayer}
		\draw (14.center) to (13);
		\draw (32.center) to (31);
	\end{pgfonlayer}
\end{tikzpicture}
 \]
expressing that $x + x = 0$ for both elements of the field $\Z_2$.
\end{example}

\begin{example}
  \label{example:polycirc-n}
  In the case $\PolyCirc_{\N}$ of the semiring of natural numbers, with the usual addition and multiplication, no extra generating morphisms or equations are actually necessary: all those appearing in Definition~\ref{definition:polycirc-s} may be derived from the cartesian distributive structure. To see why, notice that we may define each constant $s \in S$ as repeated addition:
  \[ \input{constant-s.tikz} \defeq \input{s-integer-multiply-s.tikz} \]
  where we define $\input{s-integer-multiply-n.tikz}$ inductively as
  \[
    \input{s-integer-multiply-0.tikz} \defeq \begin{tikzpicture}
	\begin{pgfonlayer}{nodelayer}
		\node [style=node] (13) at (-0.5, 0) {};
		\node [style=none] (14) at (-1.25, 0) {};
		\node [style=node] (31) at (0.5, 0) {};
		\node [style=none] (32) at (1.25, 0) {};
	\end{pgfonlayer}
	\begin{pgfonlayer}{edgelayer}
		\draw (14.center) to (13);
		\draw (32.center) to (31);
	\end{pgfonlayer}
\end{tikzpicture}

    \qquad \qquad
    \input{s-integer-multiply-n.tikz} \defeq \input{s-integer-multiply-inductive.tikz}
  \]
  The equations expressing addition and multiplication in $\N$ are then a consequence of those of cartesian distributive categories. In fact, from this observation we have that $\PolyCirc_{\N}$ is the free cartesian distributive category on one generating object.
\end{example}

\begin{example}
  \label{example:polycirc-zn}
  In a straightforward generalization of $\PolyCirc_{\Z_2}$,
  we can define $\PolyCirc_{\Z_n}$ in the same way, but with the only additional equation as
  \[ \input{s-integer-multiply-n.tikz} = \begin{tikzpicture}
	\begin{pgfonlayer}{nodelayer}
		\node [style=node] (13) at (-0.5, 0) {};
		\node [style=none] (14) at (-1.25, 0) {};
		\node [style=node] (31) at (0.5, 0) {};
		\node [style=none] (32) at (1.25, 0) {};
	\end{pgfonlayer}
	\begin{pgfonlayer}{edgelayer}
		\draw (14.center) to (13);
		\draw (32.center) to (31);
	\end{pgfonlayer}
\end{tikzpicture}
 \]
  which says algebraically that $(1 + \overset{n}{\ldots} + 1) \cdot x = n \cdot x = 0 \cdot x = 0$.
\end{example}


It is important to note that $\PolyCirc_S$ is isomorphic to the category
$\POLY_S$, defined as follows:
\begin{definition}
  $\POLY_S$ is the symmetric monoidal category with objects the natural numbers
  and arrows $m \to n$ the
  $n$-tuples of polynomials in $m$ indeterminates:
  \[ \langle p_1(\vec{x}), \ldots, p_n(\vec{x}) \rangle : m \to n \]
  with each
  \[ p_i \in S[x_1, \ldots, x_m] \]
  where $S[x_1, \ldots x_m]$ denotes the polynomial ring in $m$ indeterminates over $S$.
\end{definition}

The isomorphism $\PolyCirc_S \cong \POLY_S$ is constructed by using that
homsets $\PolyCirc_S(m, n)$ and $\POLY_S(m, n)$ have the structure of the free
module over the polynomial ring $S[x_1 \ldots x_m]^n$ which yields a unique
module isomorphism between them.
We do not prove this isomorphism here, other than to say that it follows by the
same argument as presented in \cite[Appendix A]{rdc}.

\begin{remark}
  Note in \cite{rdc} $\POLY_S$ is proven to be a reverse derivative category,
  meaning that we could have derived
  Proposition~\ref{proposition:polycirc-is-an-rdc} as a corollary of the isomorphism
  $\PolyCirc_S \cong \POLY_S$.
  We chose to provide a `native' definition of the reverse differential
  combinator of $\PolyCirc_S$ because--as we will see shortly--we will need
  to extend it with an additional generator.
  The reason for this is to gain the property of `functional completeness',
  which will allow us to express any function $S^m \to S^n$.
  This new derived category will in general no longer be isomorphic to
  $\POLY_S$, and so we must prove it too is an RDC: we do this straightforwardly
  using Theorem \ref{theorem:extension}.
\end{remark}

\begin{remark}
  When $S$ is a bonafide ring, we may account for its inverse by extending
  $\PolyCirc_S$ with a `negate' generating morphism $\begin{tikzpicture}
	\begin{pgfonlayer}{nodelayer}
		\node [style=negate] (0) at (0, 0) {};
		\node [style=none] (1) at (-1, 0) {};
		\node [style=none] (2) at (1, 0) {};
	\end{pgfonlayer}
	\begin{pgfonlayer}{edgelayer}
		\draw (1.center) to (0);
		\draw (0) to (2.center);
	\end{pgfonlayer}
\end{tikzpicture}
$,
  together with the additional equation $\input{s-negate-x.tikz} = \begin{tikzpicture}
	\begin{pgfonlayer}{nodelayer}
		\node [style=node] (13) at (-0.5, 0) {};
		\node [style=none] (14) at (-1.25, 0) {};
		\node [style=node] (31) at (0.5, 0) {};
		\node [style=none] (32) at (1.25, 0) {};
	\end{pgfonlayer}
	\begin{pgfonlayer}{edgelayer}
		\draw (14.center) to (13);
		\draw (32.center) to (31);
	\end{pgfonlayer}
\end{tikzpicture}
$.
  Then Theorem \ref{theorem:extension} suggests us how to extend the reverse
  differential combinator of $\PolyCirc_S$ to this new category:
\[ \R\left[\begin{tikzpicture}
	\begin{pgfonlayer}{nodelayer}
		\node [style=negate] (0) at (0, 0) {};
		\node [style=none] (1) at (-1, 0) {};
		\node [style=none] (2) at (1, 0) {};
	\end{pgfonlayer}
	\begin{pgfonlayer}{edgelayer}
		\draw (1.center) to (0);
		\draw (0) to (2.center);
	\end{pgfonlayer}
\end{tikzpicture}
\right] \ := \ \input{r-negate.tikz} \]
\end{remark}

\section{Functional Completeness}
\label{section:functional-completeness}
We are now ready to consider the \emph{expressivity} of the model class of
polynomial circuits.
More concretely, for a given commutative semiring $S$,
we would like to be able to represent any function
between sets $S^m \to S^n$ as a string diagram in $\PolyCirc_S$.
This property, which we call `functional completeness', is important
for a class of machine learning models to satisfy because it guarantees that we
may always construct an appropriate model for a given dataset.
It has been studied, for instance, in the context of the various `universal approximation'
theorems for neural networks
(see e.g. \cite{universal_approximators_1}, \cite{universal_approximators_2}).

To formally define functional completeness, let us fix a finite set $S$. Recall the cartesian monoidal category $\FinSet_S$, whose objects are natural numbers and a morphism $m \to n$ is a function of type $S^m \to S^n$.

\begin{definition}
  \label{definition:functional-completeness}
  We say a category $\cat{C}$ is \deftext{functionally complete} with respect to
  a finite set $S$ when there a full identity-on-objects functor
  $\ToFinSet : \cat{C} \to \FinSet_S$.
\end{definition}

The intuition for Definition \ref{definition:functional-completeness} is that we
call a category $\cat{C}$ `functionally complete' when it suffices as a syntax
for $\FinSet_S$ --- that is, by fullness of $\ToFinSet$ we may express any morphism in $\FinSet_S$. Note however that two distinct morphisms in $\cat{C}$ may
represent the same function --- $\ToFinSet$ is not necessarily faithful.

In general, $\PolyCirc_S$ is not functionally complete with respect to $S$.
Take for example the boolean semiring $\set{B}$ with multiplication and addition
as $\mathsf{AND}$ and $\mathsf{OR}$ respectively.
It is well known~\cite{complete_logic}
that one cannot construct every function of type $\set{B}^m \to \set{B}^n$
from only these operations.

Nonetheless, we claim that in order to make $\PolyCirc_S$ functionally complete it suffices to add to its presentation just one missing ingredient: the `comparator' operation, which represents the following function:
\[
	\compare(x, y) =
    \begin{cases}
        1 & \text{if } x = y \\
        0 & \text{otherwise}
    \end{cases}
\]
The following result clarifies the special role played by the comparator.

\begin{theorem}\label{theorem:funcomp}
  Let $S$ be a finite commutative semiring.
  A category $\cat{C}$ is functionally complete with respect to $S$ iff.
  there is a monoidal functor
  $\ToFinSet : \cat{C} \to \FinSet_S$
  in whose image are the following functions:
  \begin{itemize}
    \item $\langle      \rangle \mapsto s$ for each $s \in S$ (constants)
    \item $\langle x, y \rangle \mapsto x + y$ (addition)
    \item $\langle x, y \rangle \mapsto x \cdot y$ (multiplication)
    \item $\compare$
  \end{itemize}
\end{theorem}

\begin{proof}
  Suppose $\cat{C}$ is functionally complete with respect to $S$,
  where $S$ is a finite commutative semiring.
  Then by definition there is a functor
  $\ToFinSet : \cat{C} \to \FinSet_S$
  with each of the required functions in its image.

  Now in the reverse direction, we will show that any function can be
  constructed only from constants, addition, multiplication, and comparison.
  The idea is that because $S$ is finite, we can simply encode the function table of
  any function $f : S^m \to S$ as the following expression:
  \begin{equation}
    \label{equation:function-table}
    x \mapsto \sum_{s \in S^m} \compare(s, x) \cdot f(s)
  \end{equation}
  Further, since $\cat{C}$ is cartesian, we may decompose any function $f : S^m
  \to S^n$ into an $n$-tuple of functions of type $S^m \to S$.
  More intuitively, for each of the $n$ outputs, we simply look up the
  appropriate output in the encoded function table.
\end{proof}

It follows immediately that $\PolyCirc_S$ is functionally complete with respect
to $S$ if and only if one can construct the $\compare$ function in terms of constants,
additions, and multiplications.
We illustrate one such case below.

\begin{example}
  \label{example:polycirc-zp}
  $\PolyCirc_{\Z_p}$ is functionally complete for prime $p$.
  To see why, recall Fermat's Little Theorem~\cite{fermat}, which states that
  \[ a^{p-1} \equiv 1 \text{(mod $p$)} \]
  for all $a > 0$.
  Consequently, we have that
  \[
    (p - 1) \cdot a^{p-1} + 1 =
    \begin{cases}
      1 & \text{if } a = 0 \\
      0 & \text{otherwise}
    \end{cases}
  \]
  We denote this function as $\delta(a) \defeq (p - 1) \cdot a^{p-1} + 1$
  to evoke the dirac delta `zero indicator' function.
  To construct the $\compare$ function is now straightforward:
  \[
    \compare(x_1, x_2) = \sum_{s \in S} \delta(x_1 + s) \cdot \delta(x_2 + s)
  \]
\end{example}

However, as we already observed, it is not possible in general to construct the
$\compare$ function in terms of multiplication and addition.
Therefore, to guarantee functional completeness we must \emph{extend} the
category of polynomial circuits with an additional comparison operation.

\begin{definition}
  We define by $\PolyCircEq_S$ as the cartesian distributive category presented by
  the same objects, operations, and equations of $\PolyCirc_S$,
  with the addition of a `comparator' operation
  \begin{equation}
    \label{equation:g-compare}
    \input{g-compare.tikz}
  \end{equation}
  and equations
  \begin{align}
    \label{equation:compare}
    \input{s-compare-s-s.tikz} = \begin{tikzpicture}
	\begin{pgfonlayer}{nodelayer}
		\node [style=edge] (0) at (0, 0) {};
		\node [style=none] (1) at (0.75, 0) {};
	\end{pgfonlayer}
	\begin{pgfonlayer}{edgelayer}
		\draw (1.center) to (0);
	\end{pgfonlayer}
\end{tikzpicture}

    \qquad
    \input{s-compare-s-t.tikz} = \begin{tikzpicture}
	\begin{pgfonlayer}{nodelayer}
		\node [style=node] (0) at (0, 0) {};
		\node [style=none] (1) at (0.75, 0) {};
	\end{pgfonlayer}
	\begin{pgfonlayer}{edgelayer}
		\draw (1.center) to (0);
	\end{pgfonlayer}
\end{tikzpicture}

  \end{align}
  for $s, t \in S$ with $s \neq t$.
\end{definition}

To make $\PolyCircEq_S$ a reverse derivative category, we can once again
appeal to Theorem \ref{theorem:extension}.
However, we must choose an apropriate definition of $\R[\compare]$
which is well-defined and satisfies axioms ARD.1-4.

A suggestion for this choice comes from the machine learning literature.
In particular, the use of the `straight-through' estimator in quantized neural
networks, as in e.g. \cite{bengio2013estimating}.
Typically, these networks make use of the dirac delta function in the forward
pass, but this causes a catastrophic loss of gradient information in the
backwards pass since the gradient is zero almost everywhere.
To fix this, one uses the \emph{straight-through estimator}, which instead
passes through gradients directly from deeper layers to shallower ones.

In terms of reverse derivatives, this amounts to setting $\R[\delta] = \R[\id]$.
Of course, we need to define $\R$ for the full comparator, not just the
zero-indicator function $\delta$, and so we make the following choice:

\begin{theorem}
  $\PolyCircEq_S$ is an RDC with $\R$ as for $\PolyCirc_S$,
  and
  \[
    \R\left[\input{g-compare.tikz}\right] \defeq \input{r-add.tikz}
  \]
\end{theorem}

\begin{proof}
  $\R$ is well-defined with respect to the equations \eqref{equation:compare}
  since both sides of each equation must equal the unique discard morphism
  $\begin{tikzpicture}
	\begin{pgfonlayer}{nodelayer}
		\node [style=node] (0) at (0, 0) {};
		\node [style=none] (1) at (-0.75, 0) {};
	\end{pgfonlayer}
	\begin{pgfonlayer}{edgelayer}
		\draw (1.center) to (0);
	\end{pgfonlayer}
\end{tikzpicture}
$.
  Further, $\Rd{g-compare}$ satisfies axioms ARD.2-4 in the same way that $\Rd{g-add}$
  does, and so by Theorem~\ref{theorem:extension} $\PolyCircEq_S$ is a reverse
  derivative category.
\end{proof}

From Theorem~\ref{theorem:funcomp}, we may derive:
\begin{corollary}\label{cor:polycircEq_funcomp}
	$\PolyCircEq_S$ is functionally complete with respect to $S$.
\end{corollary}

Finally, note that we recover the dirac delta function by `capping' one of the
comparator's inputs with the zero constant:
\[ \delta \defeq \input{s-compare-zero.tikz} \]
whose reverse derivative is equivalent to the `straight-through' estimator:
\[ \R\left[ \input{s-compare-zero.tikz} \right] = \begin{tikzpicture}
	\begin{pgfonlayer}{nodelayer}
		\node [style=node] (13) at (0.5, 0.25) {};
		\node [style=none] (43) at (-0.5, 0.25) {};
		\node [style=none] (44) at (-0.5, -0.25) {};
		\node [style=none] (45) at (1, -0.25) {};
	\end{pgfonlayer}
	\begin{pgfonlayer}{edgelayer}
		\draw (43.center) to (13);
		\draw (44.center) to (45.center);
	\end{pgfonlayer}
\end{tikzpicture}
 = \R\left[ \begin{tikzpicture}
	\begin{pgfonlayer}{nodelayer}
		\node [style=none] (1) at (0.5, 0) {};
		\node [style=none] (2) at (-0.5, 0) {};
	\end{pgfonlayer}
	\begin{pgfonlayer}{edgelayer}
		\draw (2.center) to (1.center);
	\end{pgfonlayer}
\end{tikzpicture}
 \right] \]


\section{Polynomial Circuits in Machine Learning: Case Studies}
\label{section:saturating-arithmetic}
We now discuss the implications of some specific choices of semiring from a machine learning perspective.
Let us begin with two extremes: neural networks, and the boolean circuit models
of~\cite{rda}.

\paragraph{Neural Networks}
We may think of a neural network as a circuit whose wires
carry values in $\mathbb{R}$.
Of course, in order to compute with such circuits we must make a finite
approximation of the reals--typically using floating-point numbers.
However, this approximation introduces two key issues.
First, floating point arithmetic is significantly slower than integer arithmetic.
Second, the floating point operations of addition and multiplication are not
even associative, which introduces problems of \emph{numerical instability}.
Although attempts exist to address issues of floating point arithmetic (such as
`posits'~\cite{posit}), these still do not satisfy the ring axioms;
to properly account for these approximations would require additional work.

\paragraph{Boolean Circuits and $\Z_2$}
One may note that since we must always eventually deal with finite
representations of values, we may as well attempt to define our model class directly
in terms of them.
This is essentially the idea of~\cite{rda}: the authors use the category
$\PolyCirc_{\Z_2}$ (which they call simply $\mathbf{PolyCirc}$) as a model class
since it is already functionally complete\footnote{We discuss why in Example
\ref{example:polycirc-zp}} and admits a reverse derivative operator.
However, using a semiring of modular arithmetic in general introduces a
different problem: one must be careful to construct models so that gradients do
not `wrap around'.
Consider for example the model below, which can be thought of as two independent
sub-models $f_1$ and $f_2$ using the same parameters\footnote{
  This approach is called `weight-tying' in neural networks literature.
} but applied to different parts of the input $X_1$ and $X_2$
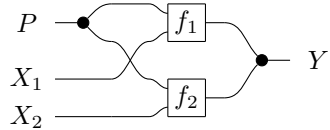
\[ \input{ex-model-sum.tikz} \]
Since $\Rd{g-copy} = \Rd{r-copy}$, when we compute the gradient update for $P$
we will sum the gradients of $f_1$ and $f_2$.
In the extreme case when the underlying semiring is $\Z_2$, then when the
gradients of $f_1$ and $f_2$ are both $1$, the result will `wrap around' to $0$
and $P$ will not be updated.
This is clearly undesirable: here we should prefer that $1 + 1 = 1$ to $1 + 1 = 0$.

\paragraph{Saturating Arithmetic}
Another possible solution is to use the semiring $\Sat_n$ as a model of
\emph{saturating unsigned integer arithmetic} for a given `precision' $n$.
The underlying set is simply the finite set $\fin{n}$, with addition and
multiplication defined as for the naturals, but `truncated' to at most $n-1$.
We define $\Sat_n$ as follows, noting that it is equivalent to the semiring
$B(n, n - 1)$ first defined in \cite[Example 3]{commutative_semirings_and_lattices}
(see also \cite{semirings_and_applications}).

\begin{definition}
  The semiring $\Sat_n$ has as addition and multiplication the operations
  \[
    x_1 + x_2 \defeq \min(n-1, x_1 + x_2) \qquad x_1 \cdot x_2 \defeq \min(n-1, x_1 \cdot x_2)
  \]
  over the set $\fin{n} \defeq \{ 0 \ldots n - 1 \}$
\end{definition}

Note that while $\Sat_n$ is a commutative semiring, it is certainly \emph{not} a ring:
the introduction of inverses means that the associativity axiom of semirings is violated.

Finally, note that for each of these choices of semiring $S$,
in general $\PolyCirc_S$ is not functionally complete.
Thus, in order to obtain a model class which is functionally complete and is a
reverse derivative category, we must use $\PolyCircEq_S$.

\section{Conclusions and Future Work}
\label{section:conclusions}
In this paper, we studied in terms of algebraic presentations categories of polynomial circuits, whose reverse derivative structure makes them suitable for machine learning.
Further, we showed how this class of categories is functionally complete for finite number
representations, and therefore provides sufficient expressiveness.
There remain however a number of opportunities for theoretical and empirical
work.

On the empirical side, we plan to use this work combined with
data structures and algorithms like that of \cite{cost_of_compositionality}
as the basis for practical machine learning tools.
Using these tools, we would like to experimentally verify that models built
using semirings like those presented in
Section~\ref{section:saturating-arithmetic}
can indeed be used to develop novel model architectures for benchmark datasets.

There also remains a number of theoretical avenues for research.
First, we want to generalise our approach to functional completeness to the
continuous case, and then to more abstract cases such as polynomial circuits
over the Burnside semiring.
Second, we want to extend the developments of Section~\ref{section:extension-theorem} in order to provide a reverse derivative structure
for circuits with notions of feedback and delay, such as the stream functions
described in~\cite{full_abstraction_digital_circuits}.


\bibliographystyle{splncs04}
\bibliography{main}

\newpage

\appendix
\section{Graphical Proofs of Extension Theorem}
\label{section:appendix-extension-theorem}
We now prove Theorem \ref{theorem:extension}.
We split the proof into the following lemmas:
\begin{enumerate}
  \item ARD.2 is preserved by composition
  \item ARD.2 is preserved by tensor product
  \item ARD.3/RD.6 is preserved by composition
  \item ARD.3/RD.6 is preserved by tensor product
  \item ARD.4/RD.7 is preserved by composition
  \item ARD.4/RD.7 is preserved by tensor product
\end{enumerate}
In each case, when we say `ARD.x is preserved by composition' we mean that if
$f$ and and $g$ satisfy ARD.x, then so too does $f \cp g$, and likewise for
tensor product.
Let us now address these lemmas in order.

\begin{lemma} ARD.2 is preserved by composition
\end{lemma}

\begin{proof}
  Assume that ARD.2 holds for $f : A \to B$ and $g : B \to C$.
  For the zero case, apply the chain rule and use the hypothesis twice to obtain the
  result as follows:
  \begin{align*}
    \input{proof-composition-preserves-rd2-z-lhs.tikz}
      & = \input{proof-composition-preserves-rd2-z-0.tikz} \\ \\
      & = \input{proof-composition-preserves-rd2-z-1.tikz} \\ \\
      & = \input{proof-composition-preserves-rd2-z-2.tikz} \\ \\
      & = 
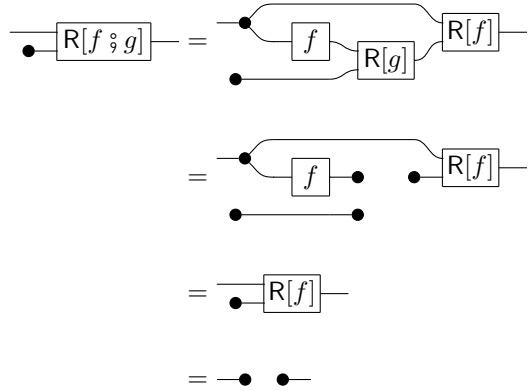
\begin{tikzpicture}
	\begin{pgfonlayer}{nodelayer}
		\node [style=none] (3) at (-1.25, 0) {};
		\node [style=node] (26) at (0.5, 0) {};
		\node [style=none] (30) at (1.25, 0) {};
		\node [style=node] (37) at (-0.5, 0) {};
	\end{pgfonlayer}
	\begin{pgfonlayer}{edgelayer}
		\draw (26) to (30.center);
		\draw (3.center) to (37);
	\end{pgfonlayer}
\end{tikzpicture}
 \\ \\
  \end{align*}
  In the additive case we proceed similarly by expanding definitions, applying
  the hypothesis, and then using associativity and commutativity of
  $\generator{g-copy}$ to obtain the final result:
  \begin{align*}
    \input{proof-composition-preserves-rd2-a-lhs.tikz}
      & = \input{proof-composition-preserves-rd2-a-0.tikz} \\ \\
      & = \input{proof-composition-preserves-rd2-a-1.tikz} \\ \\
      & = \input{proof-composition-preserves-rd2-a-2.tikz} \\ \\
      & = \input{proof-composition-preserves-rd2-a-3.tikz} \\ \\
      & = \input{proof-composition-preserves-rd2-a-4.tikz} \\ \\
      & = \input{proof-composition-preserves-rd2-a-5.tikz} \\ \\
  \end{align*}
\end{proof}

\begin{lemma} ARD.2 (RD.2) is preserved by tensr product
  For the zero case,
  \begin{align*}
    \input{proof-tensor-preserves-rd2-z-lhs.tikz}
      = \input{proof-tensor-preserves-rd2-z-0.tikz}
      = \input{proof-tensor-preserves-rd2-z-1.tikz}
      = \input{proof-tensor-preserves-rd2-z-2.tikz}
  \end{align*}
  And now in the additive case,
  \begin{align*}
    \input{proof-tensor-preserves-rd2-a-lhs.tikz}
      & = \input{proof-tensor-preserves-rd2-a-0.tikz} \\ \\
      & = \input{proof-tensor-preserves-rd2-a-1.tikz} \\ \\
      & = \input{proof-tensor-preserves-rd2-a-2.tikz} \\ \\
      & = \input{proof-tensor-preserves-rd2-a-3.tikz} \\ \\
      & = \input{proof-tensor-preserves-rd2-a-4.tikz} \\ \\
  \end{align*}
\end{lemma}

\begin{lemma} ARD.3 (RD.6) is preserved by composition
\end{lemma}

\begin{proof}
  Assume that ARD.3 holds for $f : A \to B$ and $g : B \to C$. Now calculate:
  \begin{align*}
    \D_C[\R[f \cp g]]
    & = \input{proof-composition-preserves-rd6-0.tikz} \\ \\
    & = \input{proof-composition-preserves-rd6-1.tikz} \\ \\
    & = \input{proof-composition-preserves-rd6-2.tikz} \\ \\
    & = \input{proof-composition-preserves-rd6-3.tikz} \\ \\
    & = \input{proof-composition-preserves-rd6-4.tikz}
  \end{align*}
\end{proof}

Note that in the first step, we must expand $\R^{(3)}$ using repeated
application of the chain rule- we have omitted much of this tedious calculation.
In the second step where we apply the inductive hypothesis, then
naturality of $\generator{g-discard}$ to finally obtain the result.

\begin{lemma} ARD.3 (RD.6) is preserved by tensor product
\end{lemma}

\begin{proof}
  Assume that ARD.3 holds for $f$ and $g$.
  Then it holds for $f \otimes g$ as follows:
  \begin{align*}
    \input{proof-tensor-preserves-rd6-lhs.tikz}
    & = \input{proof-tensor-preserves-rd6-0.tikz} \\ \\
    & = \input{proof-tensor-preserves-rd6-1.tikz} \\ \\
    & = \input{proof-tensor-preserves-rd6-rhs.tikz} \\ \\
  \end{align*}
\end{proof}

\begin{lemma} ARD.4 (RD.7) is preserved by composition
\end{lemma}

\begin{proof}
  Assume that ARD.4 holds for $f : A \to B$ and $g : B \to C$.
  Now we can calculate as follows:
  \begin{align*}
    \D^{(2)}[f \cp g]
    & = \scalebox{0.8}{\input{proof-composition-preserves-rd7-1.tikz}} \\ \\
    & = \scalebox{0.8}{\input{proof-composition-preserves-rd7-2.tikz}} \\ \\
    & = \scalebox{0.8}{\input{proof-composition-preserves-rd7-3.tikz}} \\ \\
    & = \scalebox{0.8}{\input{proof-composition-preserves-rd7-4.tikz}} \\ \\
    & = \scalebox{0.8}{\input{proof-composition-preserves-rd7-rhs.tikz}}
  \end{align*}
  As with the proof for RD.6 we omit a great deal of tedious expansion and
  calculation from the first step of the proof, which is obtained simply by
  expanding $\D^{(2)}[f \cp g]$ in terms of $\R$ and using naturality of
  $\generator{g-discard}$ to simplify the result.
  In remaining steps, we apply of the assumption that ARD.4 holds for $f$ and
  $g$, before finally using naturality of $\generator{g-copy}$.
\end{proof}

Note that although the above derivation is written in terms of $\D$, each step
of the proof treats the $\D$ operator merely as a syntactic sugar for its
definition in terms of $\R$.
Each step of the proof thus uses only axioms of RDCs, rather than the forward
differential structure defined in terms of it.

\begin{lemma} ARD.4 (RD.7) is preserved by tensor product
\end{lemma}

\begin{proof}
  Assume that ARD.4 holds for $f : A_1 \to B_1$ and $g : A_2 \to B_2$.
  Then we may calculate as follows, first expanding the definition of $\D$,
  and then using the inductive hypothesis to obtain the result:
  \begin{align*}
    \input{proof-tensor-preserves-rd7-lhs.tikz}
    & = \input{proof-tensor-preserves-rd7-0.tikz} \\ \\
    & = \input{proof-tensor-preserves-rd7-1.tikz} \\ \\
    & = \input{proof-tensor-preserves-rd7-2.tikz} \\ \\
    & = \input{proof-tensor-preserves-rd7-rhs.tikz} \\ \\
  \end{align*}
\end{proof}

It is now straightforward to prove Theorem \ref{theorem:extension}.

\begin{proof} (Proof of Theorem \ref{theorem:extension}) \\
  Suppose $\cat{C}$ is a category presented by generators and relations which is
  equipped with a (well-defined) $\R$ operator such that axioms ARD.1-4 hold
  for each generator, and that $\R$ is defined on tensor and composition of morphisms as in ARD.1.
  By the lemmas above, composition and tensor product preserve the remaining axioms ARD.2-4,
  and so $\cat{C}$ is an RDC.
\end{proof}

\end{document}